\documentclass[10pt,journal,compsoc]{IEEEtran}

\usepackage{amsmath,amsfonts}
\usepackage{amssymb}
\usepackage{algorithm}
\usepackage{array}
\usepackage[caption=false,font=normalsize,labelfont=sf,textfont=sf]{subfig}
\usepackage{textcomp}
\usepackage{stfloats}
\usepackage{url}
\usepackage{verbatim}
\usepackage{graphicx}
\usepackage{cite}
\usepackage{booktabs}       
\usepackage{nicefrac}       
\usepackage{microtype}      
\usepackage{xcolor}         
\usepackage{graphicx}
\usepackage{float} 

\usepackage{algpseudocode}

\usepackage{multirow}

\usepackage{adjustbox}

\usepackage{pifont}

\usepackage{authblk}

\newcommand{\taskname} {ADI}
\newcommand{\datasetname} {RescueADI}

\newcommand{\citep}[1]{{\cite{#1}}}
\newcommand{\citet}[1]{{\cite{#1}}}

\newcommand{\todo}[1]{{} }

\newcommand{\topone}[1]{{\textbf{#1}} }

\begin{document}

\title{RescueADI: Adaptive Disaster Interpretation in Remote Sensing Images with Autonomous Agents}

\author[1]{Zhuoran Liu}
\author[1,2]{Danpei Zhao*}
\author[1,2]{Bo Yuan}

\affil[1]{Image Processing Center, Beihang University, Beijing 102206, China}
\affil[2]{Tianmushan Laboratory, Hangzhou 311115, China}



\maketitle

\begin{abstract}

Current methods for disaster scene interpretation in remote sensing images~(RSIs) mostly focus on isolated tasks such as segmentation, detection, or visual question-answering~(VQA). However, current interpretation methods often fail at tasks that require the combination of multiple perception methods and specialized tools. To fill this gap, this paper introduces Adaptive Disaster Interpretation (\taskname{}), a novel task designed to solve requests by planning and executing multiple sequentially correlative interpretation tasks to provide a comprehensive analysis of disaster scenes. To facilitate research and application in this area, we present a new dataset named \datasetname{}, which contains high-resolution RSIs with annotations for three connected aspects: planning, perception, and recognition. The dataset includes 4,044 RSIs, 16,949 semantic masks, 14,483 object bounding boxes, and 13,424 interpretation requests across nine challenging request types. Moreover, we propose a new disaster interpretation method employing autonomous agents driven by large language models (LLMs) for task planning and execution, proving its efficacy in handling complex disaster interpretations. The proposed agent-based method solves various complex interpretation requests such as counting, area calculation, and path-finding without human intervention, which traditional single-task approaches cannot handle effectively. Experimental results on \datasetname{} demonstrate the feasibility of the proposed task and show that our method achieves an accuracy 9\% higher than existing VQA methods, highlighting its advantages over conventional disaster interpretation approaches. The dataset will be publicly available.

\end{abstract}


\section{Introduction}

Disaster detection based on remote sensing images~(RSIs) provides accurate, efficient, and wide-area disaster assessment and situation judgment~\cite{earthquake_damage}. With the development of remote sensing image interpretation technology, deep learning models have played an important role in various tasks including land resource statistics~\cite{land_cover_and_crop}, urban planning~\cite{huang2018urban}, and disaster monitoring~\cite{shen2021bdanet}, etc. Especially in disaster damage assessment, neural networks can quickly extract disaster-related information from remote sensing images and provide it to rescuers for analysis to help subsequent disaster assessment and response.

In terms of task formation, most of the existing disaster monitoring models are focused on isolated tasks, such as segmentation~\cite{rahnemoonfar2023rescuenet} and scene classification~\cite{aider_dataset}. Segmentation tasks can be further divided into semantic segmentation~\cite{rahnemoonfar2023rescuenet} and change detection~\cite{gupta2019creatingxbd}. Concretely, semantic segmentation takes an input image and predicts pixel-level damage level, while change detection utilizes two images of the same location with different dates and predicts a pixel-level mask to indicate the change of semantics. Scene classification~\cite{aider_dataset} aims to categorize the type of disaster or the level of damage at the image level. These perception tasks are good at extracting information from the input image. Numerous techniques, including attention~\cite{shen2021bdanet} and pyramid pooling~\cite{Bai2020PyramidPM}, have been developed to enhance the accuracy of perception systems. However, the extracted information is not in the form of natural language and requires further recognition to provide effective guidance to the rescue missions. To cope with more flexible application scenarios, visual question-answering (VQA) has emerged and become a popular research direction for disaster assessment~\cite{vqa,Sarkar2023SAMVQASA}. VQA methods respond to user input in natural language and give answers in text form, providing highly abstract feedback and reducing the understanding cost. Recent developments in large language models (LLMs)~\cite{zhao2023surveylargelanguagemodels} and visual-language models (VLMs)~\cite{zhang2024visionlanguagemodelsvisiontasks} have also further extended the boundary of VQA models. However, the VQA task does not explicitly produce intermediate perception results and thus lacks transparency. When it comes to questions related to quantitative analysis, even the largest LLMs can not avoid the problem of hallucination~\cite{llm_hallucination}, making it difficult for them to output realistic and accurate numbers. In addition, VQA methods are still limited as the task form only supports answering questions but not responding to human instructions for executing tasks, such as performing segmentation on a given image. 

Recently, large language models~(LLMs) have emerged to approach human-level intelligence~\cite{brown2020languagemodelsfewshotlearners,touvron2023llamaopenefficientfoundation,bai2023qwentechnicalreport}. LLM-driven autonomous agents have become increasingly capable of performing complex tasks with minimal human intervention~\cite{autonomous_agents, automous_agents_survey}. The key to the flexibility of LLM-driven autonomous agents is the ability to make plans to utilize multiple tools. Similarly, to systematically survey and analyze the disaster site, it is required to perceive the remote sensing image of the disaster from multiple aspects. In practice, human experts often combine the results of several different models or run one model after another sequentially in order to get the correct results. Inspired by this, we propose Adaptive Disaster Interpretation (\taskname{}) as a new form of task where an interpretation system is required to do planning according to the user's request about the disaster scene and invoke a series of modular sub-tasks to get an accurate and detailed answer.

\begin{figure*}[htb]
\centering
\includegraphics[width=\textwidth]{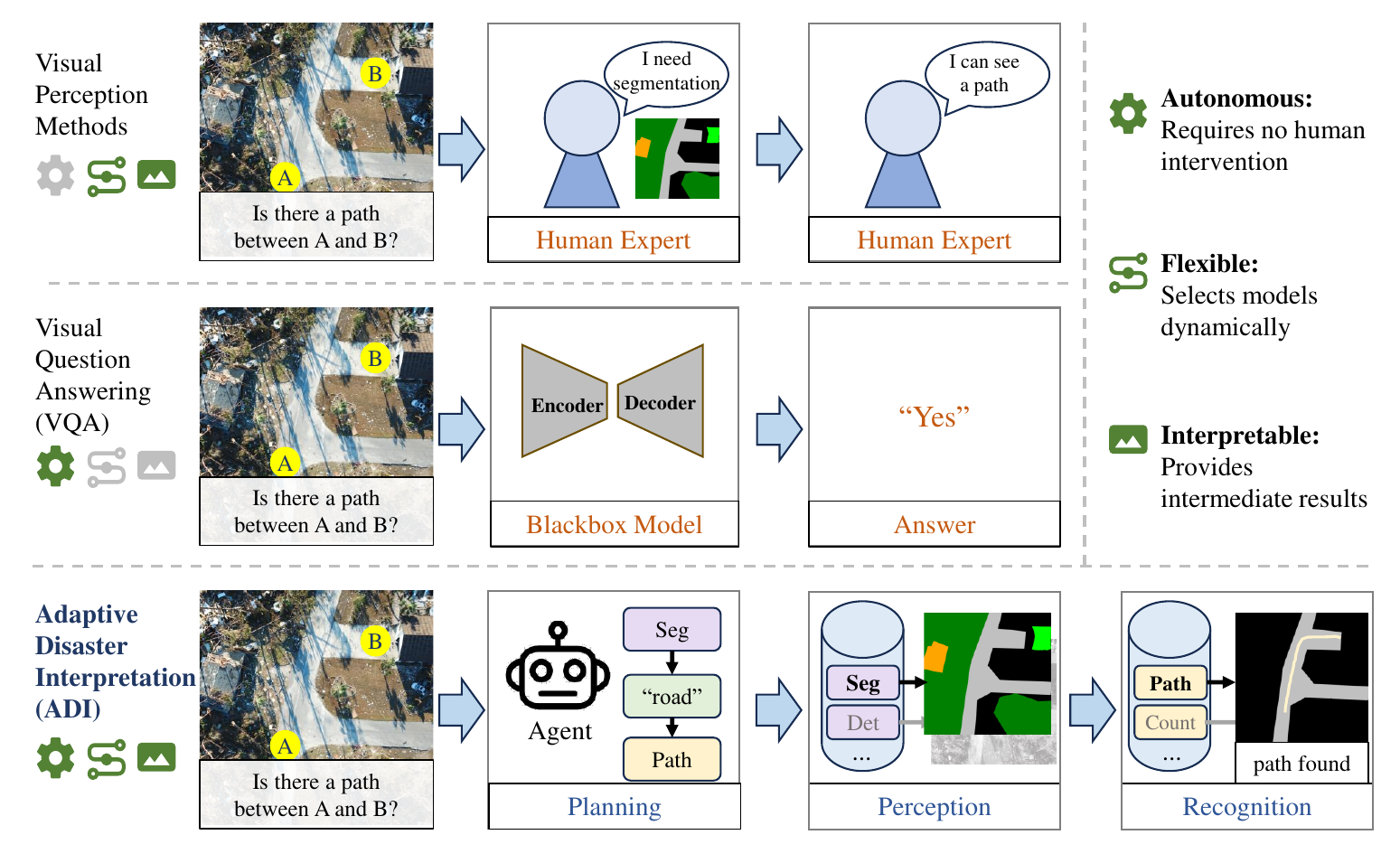}

\caption{The proposed \taskname{} integrates planning, perception, and recognition without the need of human intervention, along with clear intermediate results.}
\label{fig_task_comp_ex}
\end{figure*}

As shown in Fig.~\ref{fig_task_comp_ex}, \taskname{} unifies the need to perform individual interpretation tasks and the visual question-answering tasks that require a comprehensive understanding of the disaster scene. Compared to simply consolidating all sub-tasks into one, \taskname{} models the connections between sub-tasks through planning. Different from existing tasks, \taskname{} requires specialized interpretation of disaster scenes to be performed. For example, several interpretation tasks need to be performed sequentially to determine whether rescuers can reach the damaged houses. The first step is to determine the damage to the houses in the area as well as the obstruction of the roads, which is the basic interpretation task that current disaster detection algorithms are concerned with. Next, it is also necessary to determine whether the damaged houses can be reached, which is a high-level planning task. Finally, the results of each step are summarized into a structured text that is easy for humans to understand.

To support the research on the new task, we create \datasetname{} dataset based on high-resolution remote sensing images. Our dataset consists of 13,424 questions in 9 different aspects that are difficult to accomplish through existing single-model methods. To the best of our knowledge, this is the first disaster interpretation dataset covering multiple complex disaster interpretation scenarios for autonomous agents. Additionally, we propose an effective method to tackle the novel \taskname{} task by constructing an LLM-based autonomous agent framework, utilizing LLM's planning capabilities. Experimental results validate the feasibility of the \taskname{} task and provide a solid basis for future research.

The contribution of this paper can be summarized as follows:
\begin{itemize}
    \item We propose \taskname{}, a new task where autonomous agents use sequential modular tools to interpret complex user queries about disaster scenarios and provide more understandable responses. The task form allows adaptive interpretation based on different requests, which existing single-task frameworks can not handle.
    \item To cope with \taskname{}, a new dataset is presented named \datasetname{}. To our knowledge, this is the first time that various annotations for RSIs have been incorporated to challenge and enhance disaster interpretation tasks for autonomous agents.
    \item We introduce an autonomous agent-based framework to address the \taskname{} problem. The proposed method utilizes large language models for planning, validating the feasibility of \taskname{} for complex disaster scenarios, and achieving 9\% accuracy improvement compared to conventional VQA methods.
\end{itemize}

\section{Related Works}  \label{s_related_works}

\subsection{Rremote Sensing Interpretation in Natural Disaster Scenarios}

\begin{table*}[htb]
    \centering
    \caption{The capabilities covered by \datasetname{}.}
    \begin{tabular}{@{}lccccccc@{}}
        \toprule
        Task                       & Planning & \multicolumn{2}{c}{Perception}    & \multicolumn{3}{c}{Recognition} \\ \cmidrule(lr){3-4} \cmidrule(lr){5-7}
                                    &          & Pixel-level & Instance-level        & Object Perception & Fine-Grained Damage & Rescue Path \\ \midrule
        xBD \cite{gupta2019creatingxbd}    &          & \checkmark   &      &       & \checkmark  &             \\
        FloodNet \cite{rahnemoonfar2021floodnet}  &          &  \checkmark &   & \checkmark         &                      &             \\
        ISBDA \cite{zhu2020msnetisbda}  &  & \checkmark & \checkmark &  &  &   \\ 
        RescueNet \cite{rahnemoonfar2023rescuenet}  &  & \checkmark &  &  & \checkmark &   \\ 
        RSVQA \cite{Lobry_2020rsvqa}  &  &  &  & \checkmark &  &   \\ 
        \datasetname{} (Ours)  & \checkmark & \checkmark & \checkmark & \checkmark & \checkmark &  \checkmark \\ 
       \bottomrule
    \end{tabular}
    \label{tab_dataset_comp}
\end{table*}

Currently, many datasets for disaster scenarios are available in the form of basic computer vision tasks such as semantic segmentation, instance segmentation, change detection, and scene classification.

Most commonly, detecting damaged areas on post-disaster images can be viewed as a segmentation task, where class labels are assigned to each pixel of the image. Semantic segmentation is a well-studied task in general remote sensing images, with many datasets available~\cite{remote_sensing_semantic_segmentaion,wang2022lovedaremotesensinglandcover,panopticperception}. Different from general semantic segmentation datasets, a disaster detection dataset uses disaster scenarios as a source of images and provides more detailed annotations of the affected area. RescueNet~\cite{rahnemoonfar2023rescuenet} provides a detailed post-disaster imagery dataset captured following Hurricane Michael, with a focus on semantic segmentation. It includes comprehensive pixel-level annotations across various classes including fine-grained building damage levels, roads, water, and vegetation. The ISBDA~\cite{zhu2020msnetisbda} dataset provides instance-level building damage masks within user-generated aerial videos from social media. This dataset focuses on quantitative model evaluation, offering new perspectives in the application of aerial video analysis for assessing building damages in post-disaster scenarios.

Detecting damaged areas given pre- and post-disaster images can be viewed as a change detection task. Change detection involves monitoring semantic changes in the surface over time and is commonly used in land surveys and environmental monitoring studies~\cite{Chen2020levir}. In natural disaster scenarios, the xBD dataset~\cite{gupta2019creatingxbd} offers a unique collection of both pre- and post-event satellite imagery, designed to support change detection and building damage assessment for disaster recovery efforts. 

Scene classification is also a viable option for post-disaster assessment. In scene classification datasets, a label is assigned to each image to indicate the category of the image. Unlike object classification datasets of natural images~\cite{object_detection_survey, coco_dataset, objects365, open_images_dataset_v4} that focus on the foreground object in the image, scene classification in remote sensing images takes into account the background information. Most scene classification datasets for RSI focus on normal scenarios~\cite{scene_classfication_capsnet, scene_classification_better_exploiting_cnn, scene_classification_sota}, while AIDER~\cite{aider_dataset}, a dataset specially crafted for various disaster scenarios, provides invaluable insights into disaster response and recovery efforts.

Despite the rapid development of RSI datasets in disaster scenarios, the above datasets are designed for specific tasks and lack flexibility. Additional work by experts is still needed to perform these tasks and convert the output of each task to information that can guide post-disaster relief efforts. Therefore, a more integrated system that can make task execution plans and synthesize information from different tasks is crucial for enhancing decision-making in post-disaster situations.

\begin{figure*}[htb]
\centering
\includegraphics[width=\textwidth]{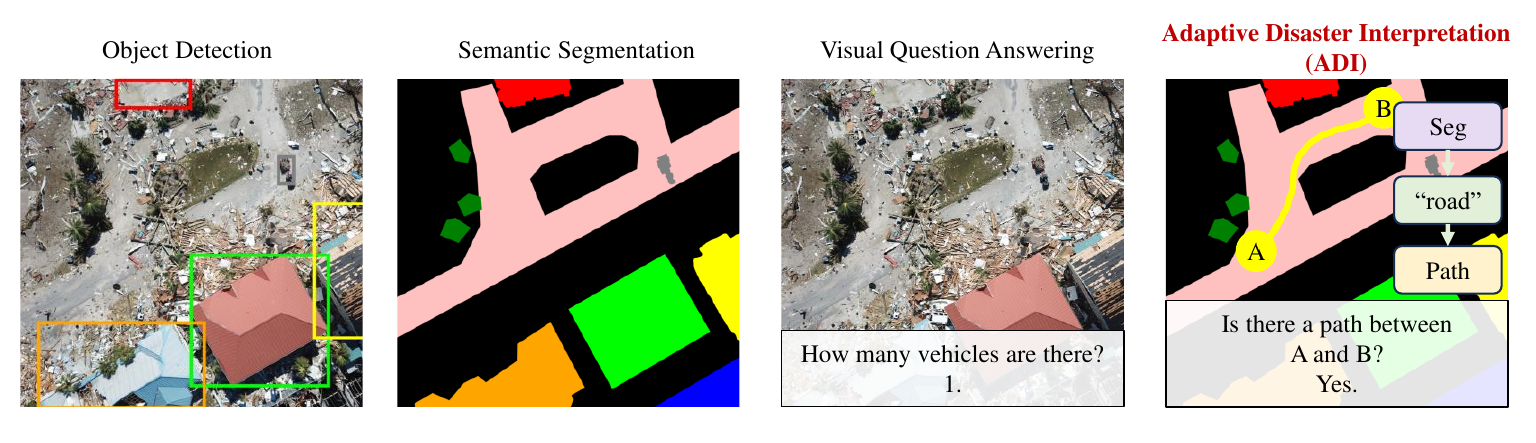}

\caption{Comparison between \taskname{} and existing task forms. Instead of performing an isolated task, \taskname{} models the connection between planning, perception and recognition.}
\label{fig_task_comp}
\end{figure*}

\subsection{VQA for Remote Sensing}

Visual Question Answering (VQA)~\cite{vqa, vqa_pretrain, vqa_okvqa, vqa_multiscale_feature_extraction, vqa_rankdvqa} is a challenging task that aims to bridge the gap between visual information and natural language understanding. Specifically, the objective of VQA is to answer questions about visual content in images with precise textual responses. With the development of LLMs, researchers have combined vision models with language models to construct vision-language models (VLMs) such as VisualGLM~\cite{ding2021cogview_visualglm} and LLaVA~\cite{liu2023llava}, which surpass the performance of traditional VQA models in various datasets. Efforts have also been made to adapt VQA to remote-sensing images. RSVQA~\cite{Lobry_2020rsvqa} proposes a system and a corresponding dataset to extract textual answers to given questions from RSI. Prompt-RSVQA~\cite{prompt_rsvqa} enhances the answer accuracy by providing a language model with contextual prompts. GeoChat~\cite{kuckreja2023geochatgroundedlargevisionlanguage} finetunes the LLaVA~\cite{liu2023llava} to answer questions on remote sensing images. As for disaster detection, FloodNet~\cite{rahnemoonfar2021floodnet} and Floodnet+~\cite{floodnetplus} combine the semantic segmentation task, the scene classification task, and the VQA task into one dataset, providing a more comprehensive understanding of disaster scenes. The dataset provides multiple question-answer pairs along with a segmentation map for each image. However, there is still a lack of modeling of the relationship between tasks and questions. These tasks are considered independent parts and are treated separately. In addition, predicting the answers to numerical questions in RSI is still a difficult task for end-to-end models as they do not explicitly perform basic perception tasks.

When human experts solve problems in disaster scenarios in remote sensing, not only do they answer questions, but they also make different plans to perform sub-tasks depending on the nature of the problem. Motivated by this, we extend the VQA framework from simply answering questions to planning sub-tasks to solve problems. In this paper, we establish \taskname{}, a new task form with the ability of planning and question answering, along with a novel dataset to support our research.

\subsection{Action Planning with Autonomous Agents}

An autonomous agent refers to a program that is able to interact with the environment according to certain rules or policy functions and accomplish certain tasks~\cite{multi_agents_survey, automatic_agent_autoact}. The recent development of LLMs has given new insights into the study of autonomous agents. Pre-trained on a large amount of text, an LLM learns a large number of logical rules and common sense and is able to drive an agent to carry out more complex task planning and external interactions~\cite{automous_agents_survey}. Camel~\cite{camel} proposes that allowing a large language model to engage in role-playing and multi-agent cooperation can enhance its ability to solve complex problems. ChatDev~\cite{chatdev} utilizes multiple LLMs to play the roles of various positions in a game development company, realizing automated, stable, and reliable implementation of complex development requirements. DEPS~\cite{deps} proposes an interactive task-planning framework that is able to reach difficult task goals in open-world game simulations. GITM~\cite{gitm} proposes a recursive tree-based task planning system that gradually disassembles complex tasks into simple ones. WebGPT~\cite{webgpt} uses GPT3 as a planner, which continuously interacts with the search engine and answers the user's questions. ToolFormer~\cite{toolformer} proposes external tools to be integrated into LLMs. VisualChatGPT~\cite{visual_chatgpt} employs GroundingDINO~\cite{groundingdino} and SAM~\cite{sam} as external tools to be invoked by the LLM and realizes the recognition and editing of input images according to the user's needs. HuggingGPT~\cite{hugginggpt} packages a large number of models on an open-source modeling platform as tools that can be invoked by an autonomous agent, which is capable of completing complex task processes that require the combination of multiple models.

In remote sensing, pioneering work has been done to utilize the power of LLM in image interpretation~\cite{xu2024rsagentautomatingremotesensing}. However, existing methods are limited to natural scenes and lack specialized tools for disaster interpretation.

Despite the rapid progress in research on action planning, most developments are tailored for specific applications. Limited efforts have been dedicated to disaster interpretation, which demands specialized models, tools, and expertise. In this paper, we propose a standardized formulation of \taskname{} and introduce a novel dataset to validate action planning in disaster scenarios. Additionally, we develop a baseline method that leverages the planning capabilities of large language models (LLMs) and incorporates specialized tools to enhance its adaptability for interpreting disaster scenarios in remote sensing images.

\section{Task Definition of \taskname{}}  \label{s_task_def}

\begin{figure*}[htb]
\centering
\includegraphics[width=\textwidth]{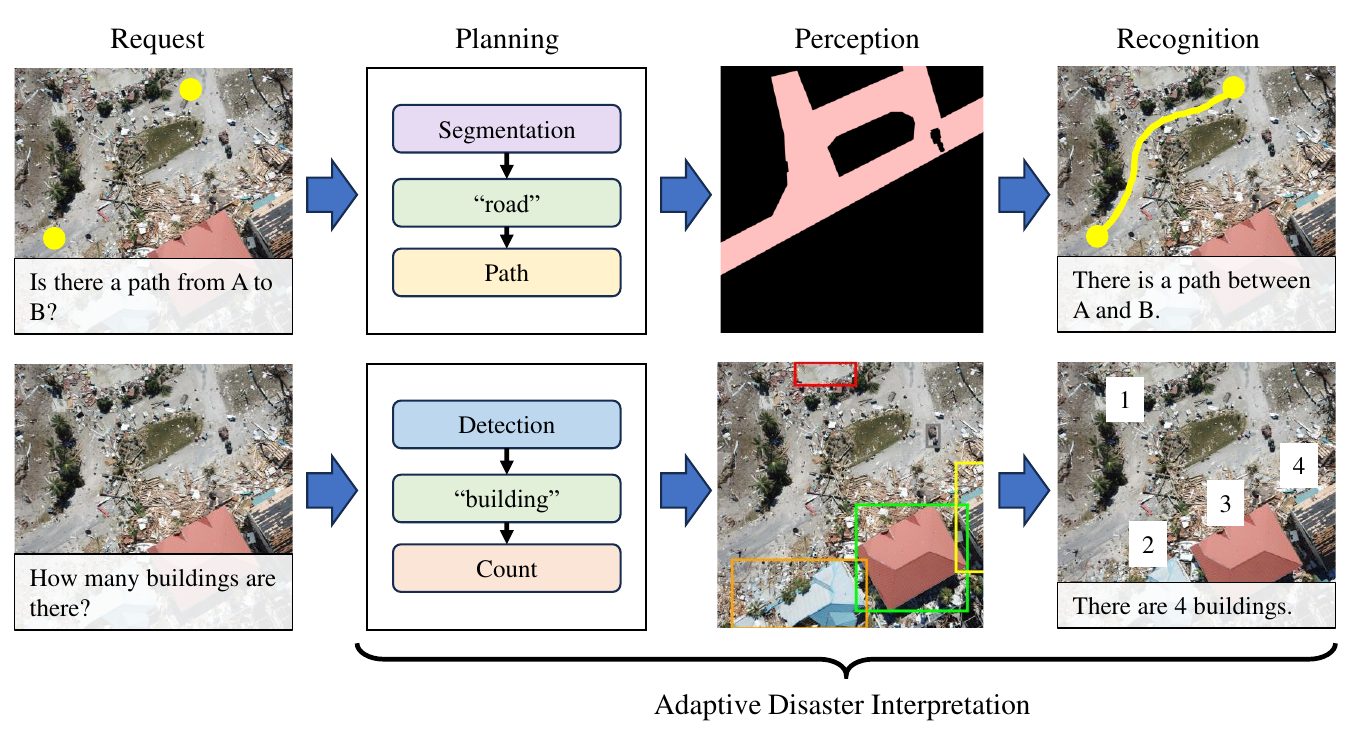}

\caption{The definition and examples of \taskname{}. The task has three stages: planning, perception, and recognition. The planning stage dynamically schedules the following perception and recognition, enabling flexible interpretation of disaster scenarios. }
\label{fig_task_demo}
\end{figure*}

In this paper, we introduce a novel task called \taskname{}, designed to enhance our understanding of disaster scenarios through an agent-based approach. As demonstrated in Fig. \ref{fig_task_comp}, our framework differs from traditional tasks by integrating sub-task planning and question-answering capabilities. Unlike conventional tasks that are limited to specific forms such as semantic segmentation or visual question answering, \taskname{} requires an agent that can dynamically adjust to the demands of the input request. For instance, it can perform detailed semantic segmentation of affected areas when required, while also being able to answer contextual questions about the disaster, such as the extent of damage or the number of affected structures. 

The input of \taskname{} consists of an input image and a textual request. The input image provides visual information about the disaster scene while the request provides the objective of the interpretation task, which could be a request to perform a specific sub-task, a question to be answered, or both. The output of \taskname{} can be defined from three connected aspects, as shown in Fig. \ref{fig_task_demo}.

1) The planning aspect asks an agent to create a plan according to the need of the input request. Each step of the plan belongs to a set of sub-tasks predefined by the dataset. The agent should perform every necessary sub-task while avoiding planning for sub-tasks that are not needed.

2) The perception aspect requires every selected sub-task to output accurate prediction results. For example, a segmentation sub-task involves assigning categories to each individual pixel within the input image, and an object detection sub-task entails the identification and localization of objects in the input image.

3) The recognition aspect demands the agent to understand and analyze the perception result and generate a direct response to the request in natural language.

Based on the above three aspects, a more specific definition of \taskname{} is presented as follows. The general input of the task is defined as a tuple $(I, Q)$, where $I \in \mathbb{R}^{C \times H \times W}$ is the input image and $Q$ is the input request in natural language. $C$ represents the number of channels in the image, $H$ is the height, and $W$ is the width of the image. The output of \taskname{} is a structured response that integrates the results from the planning, perception, and recognition aspects. Formally, the output can also be represented as a tuple $(P, R, A)$, where:

1) $P$ is a sequence of planned sub-tasks, $P = \{p_1, p_2, \ldots, p_n\}$, where each $p_i$ belongs to a predefined set of sub-tasks $\mathcal{P}$. The agent must select and plan these sub-tasks based on the input request $Q$.

2) $R$ is a set of results from the executed sub-tasks, \(R = \{r_1, r_2, \ldots, r_n\}\), corresponding to the sub-tasks in \(P\). Each \(r_i\) is the output of the sub-task \(p_i\), which could include segmentation maps, object detection bounding boxes, or other form of basic perception tasks.

3) $A$ is the final natural language answer generated in response to the input request $Q$. This answer should be coherent, contextually relevant, and derived from the request $Q$ and the perception results $R$.

The overall objective of \taskname{} is to maximize the accuracy of \(P\), \(R\), and \(A\) with respective to the input tuple \((I, Q)\).

The following metrics are employed to evaluate the performance of an agent on \taskname{}:

1) \textbf{Planning Accuracy}: To measure the correctness of the planned sub-tasks \(P\) against a ground truth sequence of sub-tasks \(P^*\), we treat the presence of each sub-task as a binary classification problem and utilize precision and recall metrics.

2) \textbf{Perception Accuracy}: Metrics commonly used in basic perception tasks are adopted to measure the perception accuracy of selected sub-tasks. For semantic segmentation sub-tasks, we use Intersection over Union (IoU) to represent the accuracy of the segmentation outputs. For object detection sub-tasks, we employ mean Average Precision (mAP) to verify the quality of the bounding boxes. 

3) \textbf{Recognition Accuracy}: Evaluate the quality of the generated answer \(A\) by comparing it to the ground truth answer \(A^*\) using exact-match accuracy and GPTScore~\cite{hugginggpt}. The former checks for identical answers, while the latter assesses semantic similarity to capture nuanced differences.

To summarize, \taskname{} represents a comprehensive and dynamic task that challenges agents to integrate planning, perception, and recognition capabilities to effectively interpret and respond to complex disaster scenarios. This approach aims to push the boundaries of current AI systems and foster advancements in disaster response technologies.

\section{Dataset}

Existing datasets for individual tasks are not capable of providing a comprehensive understanding of disaster scenes necessary to accomplish the \taskname{} task. Therefore, we construct \datasetname{}, a novel dataset designed to support the integrated task of sub-task planning, perception, and recognition. In this section, a detailed explanation of the creation of \datasetname{} is given.

\subsection{The \datasetname{} Dataset}

The proposed \datasetname{} dataset, designed for disaster scenes in remote sensing images for autonomous agents, distinguishes itself from existing datasets through its unique task format and comprehensive coverage of disaster interpretation tasks. The images of disaster scenes in our dataset are primarily sourced from the RescueNet dataset. However, we have conducted additional data cleaning and developed methods to generate high-quality request-planning-answer annotations based on the original dataset. The \datasetname{} dataset is designed to support disaster interpretation requests, focusing on six categories of semantics and four fine-grained building damage levels. Additionally, it includes nine distinct types of requests, addressing the fundamental needs and challenging scenarios encountered in remote sensing-based disaster interpretation.

\begin{figure}[htb]
\centering
\includegraphics[width=\columnwidth]{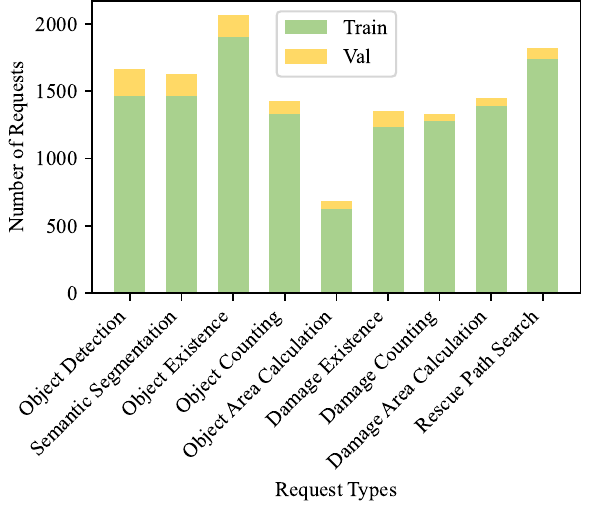}
\caption{The distribution of different request types in \datasetname{}.}
\label{fig_data_nums}
\end{figure}

\begin{figure*}[htb]
\centering
\includegraphics[width=\textwidth]{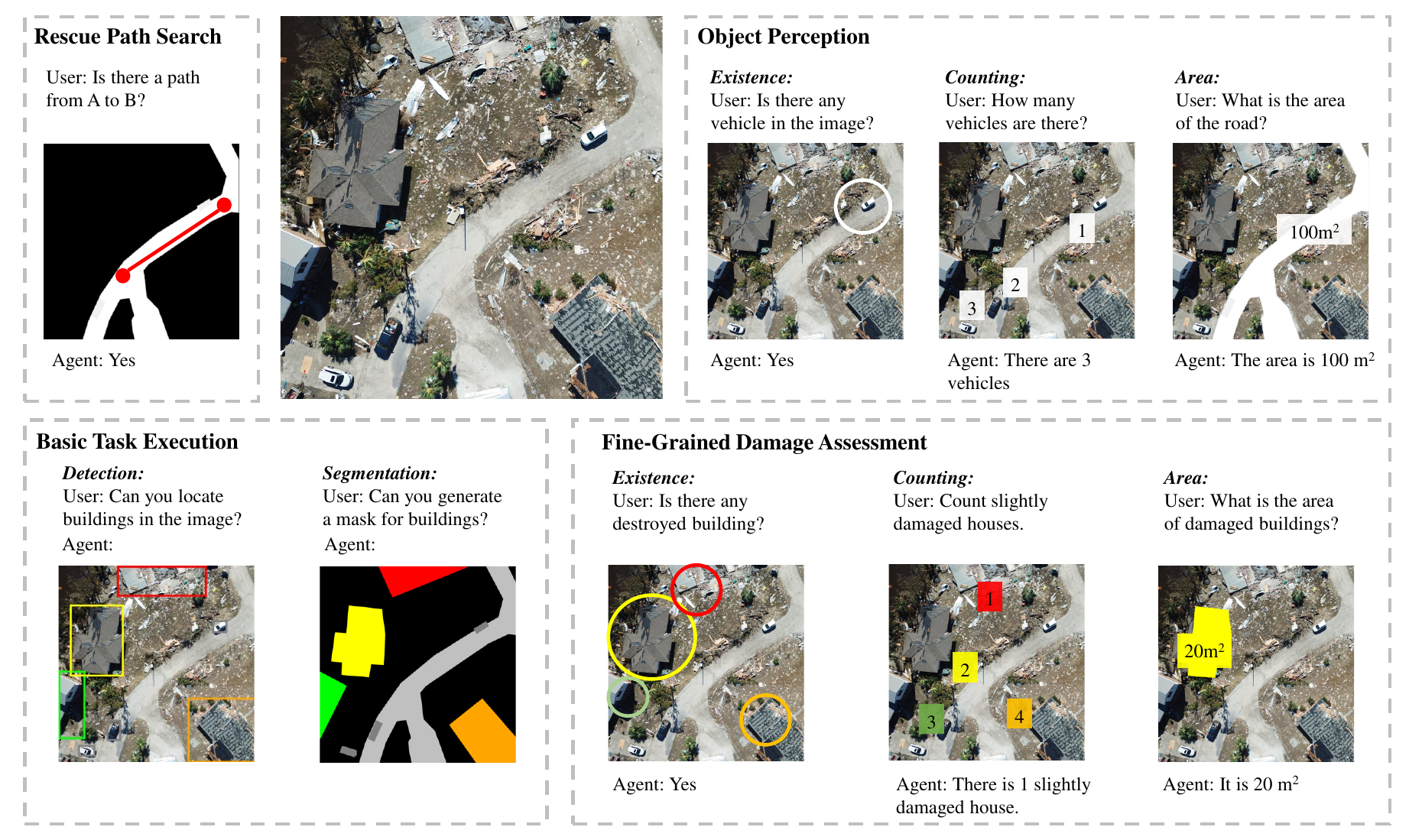}

\caption{Demonstration of different types of annotated requests in \datasetname{}. }
\label{fig_request_types}
\end{figure*}

As shown in Table. \ref{tab_dataset_comp}, compared to existing datasets, our dataset is the first one to cover the full spectrum of planning, perception, and recognition. To be specific, \datasetname{} contains 13,424 requests and answers on 4,044 remote sensing images on disaster scenes, along with 16,949 semantic masks and 14,483 object bounding boxes. The fine-grained damage levels range from no damage to totally destructed, and the categories for other semantics include water, vehicle, clear road, blocked road, pool, and trees. The types of requests comprise object detection, semantic segmentation, object existence, object counting, object area calculation, damage existence, damage counting, damage area calculation, and rescue path search. Fig. \ref{fig_data_nums} illustrates the number distribution of tasks covered by our dataset. The dataset is split into 12,426 requests for training and 998 requests for validation.

\subsection{Request Formulation}

In real-world disaster scenarios, the need for accurate and actionable information is emphasized. Responders require fast and precise data to make informed decisions that can save lives and mitigate damage. These needs can be categorized into several key areas:

\begin{itemize}
    \item Basic Task Execution: Fundamental to disaster response are tasks such as identifying critical objects (e.g., vehicles, buildings) or specific regions (e.g., blocked road and clear road) within an image. These basic identifications are crucial for initial assessments and resource allocation.

    \item Object Perception: In disaster scenes, responders need to recognize and categorize various elements within a scene. This includes tasks such as determining the existence of objects, counting the number of objects, and calculating the area they occupy. Accurate object perception aids in understanding the complexity of the scene and planning appropriate responses.

    \item Fine-Grained Damage Assessment: Detailed analysis of the extent and severity of damage is essential for prioritizing response efforts and resources. This includes tasks such as determining the existence of damage, counting instances of damage, and calculating the area affected by the damage. Fine-grained damage assessment provides a deeper insight into the disaster's impact and helps in formulating effective recovery strategies.

    \item Rescue Path Search: In chaotic and hazardous environments, identifying safe and efficient routes for rescuers is vital. This involves analyzing the terrain, debris, and potential hazards to find paths that minimize risk and maximize response efficiency. Effective rescue path search can significantly improve the speed and safety of rescue operations.
\end{itemize}

Based on these critical needs, we have constructed requests in our dataset that align with these four general types. This systematic approach is derived from a thorough analysis of real-world disaster scenarios and considers the current limitations and capabilities of existing algorithms. By addressing these specific needs, our dataset provides a comprehensive tool for improving disaster response and management.

As illustrated in Fig. \ref{fig_request_types}, nine sub-types are derived from these four general types. Basic task execution includes the demand to execute semantic segmentation or object detection on the given image. Object perception includes tasks such as determining the existence of objects, counting the objects, and calculating the area they occupy. For instance, object presence requests involve answering "yes/no" questions about the existence of certain objects, while object counting and area calculation requests require precise numerical answers. Fine-grained damage assessment involves similar tasks but focuses on damage-specific details. It includes determining the existence of damage, counting the instances of damage, and calculating the area affected by the damage. This detailed analysis helps in understanding the severity and extent of the disaster's impact. Lastly, rescue path search is crucial in disaster response as it involves discovering a safe and efficient path between given points. This task addresses the challenge of navigating complex and shifting environments, which has not been adequately covered by existing datasets.

To ensure the diversity of the request text and avoid the possibility of models overfitting to a specific input pattern, requests are created in three steps. Firstly, for each image, we generate requests of different types with a fixed template, which we refer to as the "seed request". After that, we employ the power of pretrained LLMs to expand the diversity of seed requests. For each request, an LLM is prompted to generate and randomly sample 3 different forms of the request while keeping the meaning of the request unchanged. As an LLM does not always produce accurate answers, we manually screen the dataset to filter out requests that have been rewritten in wrong ways. During the manual screening process, we carefully review each generated request to ensure that the rephrasing remains faithful to the original intent.

\subsection{Annotations}

The unique structure of the \taskname{} task necessitates a comprehensive annotation approach. Each request is annotated as a tuple of three elements \((P, R, A)\), representing planning, perception, and recognition. The planning annotation is detailed through a list of sub-tasks required to fulfill the request. The perception annotation includes a semantic segmentation map and bounding boxes for objects in the scene. The recognition annotation provides a textual response to the input request. To construct the dataset with these annotations, we employed a combination of manual annotation, automated processes, and existing datasets, ensuring that the resulting annotations are of high quality.

The planning annotation is assigned to requests based on their types during the request generation stage. Although the text of each requests are altered during the rephrasing process, the core intention remains consistent with its seed request. Consequently, the required sub-tasks are uniform across requests of the same type, and we identify these sub-tasks to assign the planning annotation accurately.

The perception annotation process includes both segmentation and object detection annotations. The RescueNet dataset provides segmentation annotations for 10 categories, including four distinct levels of building damage, meeting the requirements for segmentation tasks. However, the object detection annotation is not provided by the original dataset. To address this, we initially identify the outer bounding box for each connected component in the segmentation map of foreground objects to generate preliminary object detection annotations. These annotations are then visualized, and any inaccuracies are manually adjusted to produce accurate object detection labels for each input image. This procedure yielded 14,483 high-quality annotations of foreground objects in disaster scenes.

For the recognition annotation, as manually annotating textual answers for each image and request is expensive, we developed a process to synthesize the needed annotations from existing labels with a semi-automatic pipeline. Different methods are employed for the automated part of each request type, as detailed below.

\begin{itemize}
    \item Object Presence / Damage Level Presence: An object presence request asks to identify if there are any objects of a given class in the scene. Therefore, we utilize the segmentation label to produce the ground truth of object presence requests. The answer is "yes" if the segmentation map of a given class is not empty, and "no" otherwise.
    \item Object Counting / Damage Level Counting: A counting request requires an agent to give the exact number of objects of a given class in the scene. We use the object detection label to produce the recognition ground truth of counting requests.
    \item Area Calculation: In the area calculation task, the agent is required to estimate the area of a specified class present in the scene. To achieve this, we use the segmentation label in combination with the ground sample distance (GSD) of the original image. This allows us to calculate the actual area corresponding to the mask of the specified class.
    \item Rescue Path Search: In the rescue path search request, the agent needs to find if an unblocked route exists between given points. The start point and the destination point are manually annotated. To produce the label for this task, we utilize the segmentation label for clear roads and perform the A* path-finding algorithm to determine if the destination is reachable.
\end{itemize}

The annotation process for the \datasetname{} dataset involves a careful combination of manual and automated methods to ensure high-quality and comprehensive annotations. This multi-step approach addresses the unique requirements of planning, perception, and recognition annotations, thereby supporting the diverse needs of the \taskname{} task.

\section{Agent-based Disaster Interpretation} \label{s_methods} 

\begin{figure*}[htb]
\centering
\includegraphics[width=\textwidth]{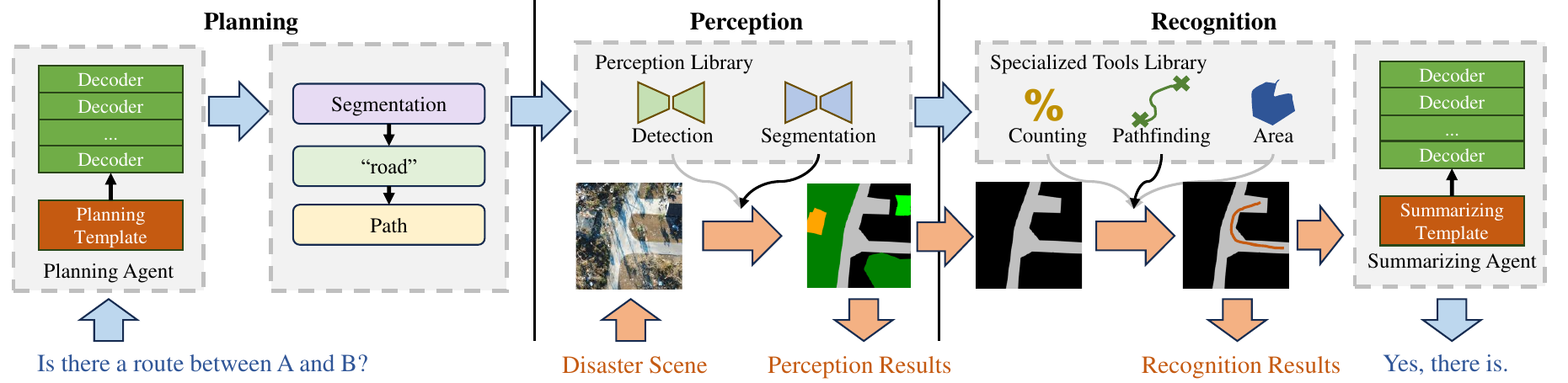}

\caption{Overall architecture of our proposed agent based \taskname{}. }
\label{fig_method}
\end{figure*}

As illustrated in Fig. \ref{fig_method}, we introduce a novel framework for agent-based \taskname{}. This approach covers planning, perception, and recognition functionalities through interactions with a set of specialized tools. The agent consists of three parts: the planner module, the perception module, and the recognition module. They are responsible for predicting the best plan to accomplish the request, providing perception results that supports the plan, and generating responses based on perception results, respectively. Different from existing end-to-end VQA models, the autonomous agent paradigm employs separate neural networks for each part of the agent, which are not connected by gradient flows during training. This design brings more efficient utilization of pre-trained large models and offers greater transparency in the decision-making process. Moreover, the use of specialized tools enhances the ability to solve numerical tasks, which are hard for traditional end-to-end models.

\subsection{Planning Module}

To instruct the LLM to output a valid plan that can be executed automatically, we construct the input prompt with the following parts:

\begin{itemize}
    \item Task Definition. The task definition asks the planning agent to create an action plan based on user requests.
    \item Format Constraints. The format constraints demonstrate the required JSON format for the LLM so that the output of the LLM can be parsed automatically. The expected format is an array of JSON objects, where each object represents a single action. An action includes the ID of the tool to use, the inputs and outputs of the tool.
    \item Tool Descriptions. The tool descriptions include a list of available tools. For each tool, a brief description is provided and the numbers and types of the inputs and outputs of the tool is given.
    \item Input Descriptions. The descriptions of the input image, including its identifier and resolution.
    \item User Request. The request is concatenated at the end of the prompt.
\end{itemize} 

The planning agent uses an LLM to process the described prompt as input and produces a formatted JSON in the form of a list of actions to execute. The LLM also assigns a unique identifier to each input and output, which is used as a key for storing intermediate results in a dictionary for later reference.

In practice, the LLM does not always generate a response in the correct format. We design a rejection sampling-based approach to handle this issue. The LLM is prompted to put the desired JSON at the end of its response. A reverse search algorithm is then used to find the valid JSON structure. The search range starts from the last character in the response string. If the substring at the end of the response is not a valid JSON, the beginning of the searching range moves backward until it reaches the start of the response. If no valid JSON is found after the search, we generate the response again with a larger temperature to get a different result until a valid JSON is found.

Once a valid action plan in JSON format is generated, we check the validity of each action in the plan. An action is considered invalid if the tool ID does not exist or the input resource ID is not available. We find that invalid actions can be classified into two categories: typographical error and hallucination. A typographical error happens when the planner agent tries to use an existing tool but generates the wrong tool ID by mistake. Hallucination, on the other hand, is when the LLM invents a tool that doesn’t exist. We identify the two categories by computing the edit distance between the LLM-generated identifier and each existing identifier. If the closest edit distance is less than 8, the identifier is corrected to the closest match. Actions that do not satisfy this criterion are classified as hallucinations and are therefore discarded.

\subsection{Perception Module}

The perception module provides tools for visual perceptions, bridging the gap between LLM and the image modality in disaster scenes. The module provides two vision actions to extract information from the input image. The detection tool takes an image as input and produces a structured representation of detected objects in JSON format. The segmentation tool takes an image as input and outputs masks of existing categories in the scene.

The object detection and semantic segmentation tasks are well-studied tasks in computer vision. Since the agent-based method has a modular structure, we can utilize existing state-of-the-art models for these tools. To implement the object detection tool, we employ the model structure of GroundingDINO~\cite{groundingdino} and train the model on the \datasetname{} dataset to obtain a standard object detection model that detects interested fine-grained categories in disaster scenes. Subsequently, we convert the detection output into a structured JSON to connect it to the agent framework. Each detected object is represented by an item in an array. An item consists of attributes describing the type of the object and the bounding box coordinates of the object.

To implement the semantic segmentation tool, we utilize the PSPNet and train it on the \datasetname{} dataset. To adapt the model into the agent framework, we train the model on \datasetname{} and convert the output into a dictionary of binary masks, where the key represents the category and the value represents the mask of the corresponding category.

\subsection{Recognition Module}

The recognition module provides specialized tools to perform recognition to the perception result and utilize an LLM-based agent to perform analysis and produce the final result.

When it comes to numerical tasks, LLMs are prone to mistakes due to the nature of token-based number prediction. To address this issue, we introduce the counting tool, which provides accurate counting results based on object detection outputs. This tool accepts the detection result and a target object type as inputs and returns the exact number of objects of that type.

For dense prediction tasks like segmentation, it is hard to directly make use of the segmentation mask with a language model. Therefore, we provide the agent with the segmentation area tool. This tool computes the total area covered by a given category from the segmentation result. It takes the segmentation map and the target category name as inputs, outputting the total area of the specified category.

To solve the path-finding problem, we integrate the A-star algorithm into the agent framework as a callable tool. The tool accepts two points and a binary mask as input and finds out if there is a path between the two points. By calling the path finding tool, the agent can get accurate results to determine if the rescuers can reach a specific destination.

After recognizing the results of the perception, a summarizing agent integrates the output of these tools into its final answer. The prompt for the summarizing agent contains the following parts:
\begin{itemize}
    \item Task Definition: Requests the agent to provide a final answer based on the user’s request.
    \item Action History: Details the sequence of actions performed and their outcomes.
    \item User Request: The original request made by the user.
\end{itemize}

Through the coordination of these modules, the agent is able to leverage specialized tools to generate accurate, transparent, and reliable outputs for disaster response tasks.

\section{Experiments} \label{s_exp} 

In this section, we conduct experiments on our \datasetname{} dataset using the proposed agent-based method to validate the effectiveness of our task. We also experiment with existing methods to showcase the unique capability of the novel task form.

\subsection{Evaluation Metrics}

The performance of methods on \taskname{} is evaluated from three aspects: planning, perception, and recognition. We evaluate the planning result with valid rate, recall, and precision. The planning agent may fail to produce a valid plan at all. Therefore, we define the valid rate as the number of valid plans divided by the total number of requests in the validation set. 

\begin{equation}
    \text{VR} = \frac{|\text{valid}|}{N}
\end{equation}

where $|\text{valid}|$ is the number of valid plans and $N$ is the number of requests. For valid plans, we define precision and recall by referring to the plan as a binary classification problem.

\begin{equation}
    \text{P} = \frac{\text{CA}}{\text{CA} + \text{UA}}
\end{equation}

\begin{equation}
    \text{R} = \frac{\text{CA}}{\text{CA} + \text{MA}}
\end{equation}

where $\text{CA}$ represents the number of correct actions predicted, $\text{UA}$ represents the number of unnecessary actions predicted, and $\text{MA}$ represents the number of missing actions, that are necessary but not predicted.

To evaluate the perception and recognition performance, we divide the requests into two groups according to their type. For requests that involve visual perception tasks, the agent outputs object detection or segmentation results, which are evaluated with mean Intersection over Union (mIoU) and mean Average Precision (mAP).

For question answering results, we adopt two different metrics to obtain a more comprehensive evaluation. The matching rate is calculated by strictly matching the ground truth with the predicted result. For area calculation requests, which require a numerical answer, we match the generated number and the ground truth. An answer is considered correct if the difference is less than one square meter or the relative difference is less than two percent. For other types of requests, the answer must match the exact ground truth to be considered correct. However, in the context of natural language, this strict matching rule may eliminate answers that are semantically correct but in a different form. Therefore, we adopt GPTScore following HuggingGPT~\cite{hugginggpt} to provide a more flexible judgment to the answers. GPTScore asks an LLM to judge the correctness of an answer given the ground truth. This metric takes into consider the semantics of the answer and produces results closer to human interpretation.

\subsection{Implementation Details}

\begin{table*}[htb]
    \centering
    \caption{Qualitative comparison between \taskname{} and existing tasks on \datasetname{}.}
    \begin{tabular}{llccccccc}
        \toprule
        Task & Model & \multicolumn{3}{c}{Planning} & \multicolumn{2}{c}{Perception} & \multicolumn{2}{c}{Recognition} \\ 
        \cmidrule(lr){3-5} \cmidrule(lr){6-7} \cmidrule(lr){8-9}
        & & VR & P & R & \(\text{mIoU}\) & \( \text{mAP}_{50}\) & Exact & GPTScore \\
        \midrule
        \multirow{2}{*}{Segmentation} 
                            & SAN &- &- &- &0.6802 &- &- &- \\
                            & PSPNet &- &- &- &0.6828 &- &- &- \\
        \midrule
        \multirow{2}{*}{Detection} 
                            & FasterRCNN &- &- &- &- &0.7640 &- &- \\
                            & GroundingDINO &- &- &- &- &0.8010 &- &- \\
        \midrule
        \multirow{2}{*}{VQA} 
                            & GeoChat &- &- &- &- &- &0.2263 &0.2975 \\
                            & VisualGLM &- &- &- &- &- &0.6535 &0.6915 \\
        \midrule
        ADI & Ours w/ PSPNet, GroundingDINO &0.9960 &0.9105 &0.9728 &0.6828 &0.8010 &\topone{0.7563}& \topone{0.7896}  \\
        \bottomrule
    \end{tabular}
    \label{tab:task_model_metrics}
\end{table*}

\begin{table*}[htb]
    \centering
    \caption{Detailed comparison of the answer accuracy by request types. In the table, \textit{dmg} denotes the accuracy of recognizing fine-grained damage levels while \textit{obj} denotes the accuracy of other objects}
    \begin{tabular}{lcccccccc}
    \toprule
                  & \multicolumn{2}{c}{existence} & \multicolumn{2}{c}{counting} & \multicolumn{2}{c}{area} & path & all      \\ \cmidrule(lr){4-5} \cmidrule(lr){6-7} \cmidrule(lr){2-3}
                  & obj      & dmg                & obj      & dmg                 & obj   & dmg           &             &          \\ \midrule
    GeoChat     &  0.4192 &0.3265 &0.0345 &0.1453 &0.2222 &0.3000 &0.4744 &0.2975    \\ 
    VisualGLM     & 0.6527 &0.6633 &0.8103 &0.7436 &0.3333 &0.3500 &0.8462 &0.6535    \\ 
    Ours     & \topone{0.8024} & \topone{0.8367} & \topone{0.8621} & \topone{0.8462} & \topone{0.7222} & \topone{0.4333} & \topone{0.8846} & \topone{0.7896}    \\ 
    \bottomrule
    \end{tabular}
    \label{tab:per_type}
\end{table*}

\begin{figure*}[htb]
\centering
\includegraphics[width=\textwidth]{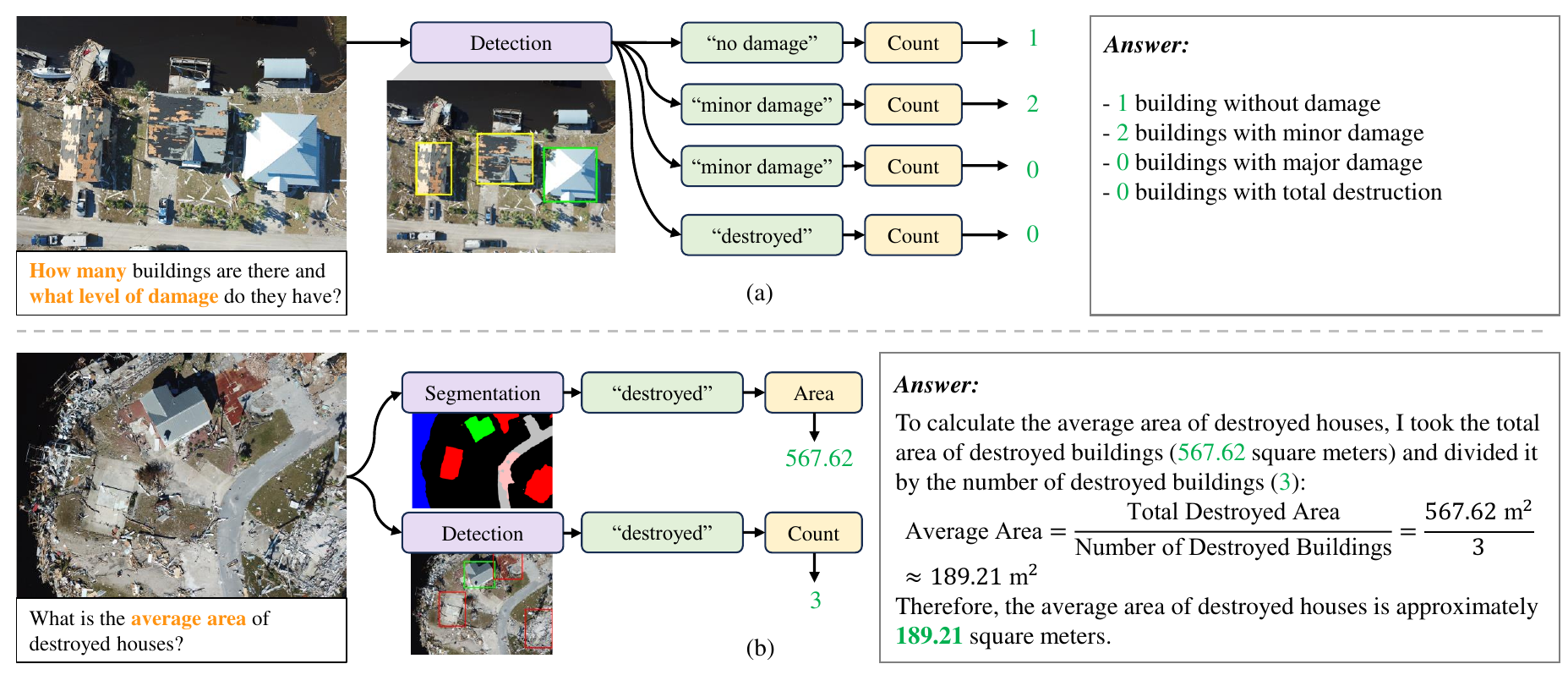}

\caption{Visualization of the planning, perception, and recognition results of the proposed method on complex requests. (a) A compound request. (b) A request that requires advanced reasoning.}
\label{fig_ood}
\end{figure*}

\subsubsection{Large Language Model}

The modular framework of our method enables us to experiment with different LLMs without the need to retrain other parts of the agent. In our experiments, we use the close-sourced GPT4o-mini as the backend of the planning and summarizing agent. In the ablation studies, we also experiment with state-of-the-art open-source LLMs. When generating the output with LLMs, we adopt the greedy decoding strategy with a temperature of 0.7.

\subsubsection{Perception Models}

The perception models are trained on four RTX3090 graphic cards. The PSPNet and FasterRCNN are initialized with pretrained ResNet-50~\cite{resnet} backbones and trained with the SGD optimizer. The SAN and GroundingDINO are initialized with pretrained ViT-B~\cite{vit} backbones and trained with the AdamW optimizer.

\begin{table*}[htb]
    \centering
    \caption{Ablation study on LLM backends.}
    \begin{tabular}{lccccc}
        \toprule
        Model & \multicolumn{3}{c}{Planning} & \multicolumn{2}{c}{Recognition} \\ 
        \cmidrule(lr){2-4} \cmidrule(lr){5-6}
        &VR & P & R & Exact & GPTScore \\
        \midrule
        Ours w/ GPT 4o mini &\topone{0.9960}& 0.9105& \topone{0.9728}& \topone{0.7563}& \topone{0.7848}  \\
        Ours w/ Deepseek 7b &0.2615& 0.5781& 0.6667& 0.0728& 0.1297  \\
        Ours w/ Qwen1.5 7b &0.6603& 0.7932& 0.8558& 0.3006& 0.3861  \\
        Ours w/ Qwen2.5 7b &0.9890& \topone{0.9573}& 0.6586& 0.5411& 0.6867  \\
        Ours w/ Llama3.1 8b &0.8537& 0.6629& 0.9554& 0.5269& 0.4937 \\
        \bottomrule
    \end{tabular}
    \label{tab:abl_llm}
\end{table*}

\begin{table*}[htb]
    \centering
    \caption{Ablation study on specialized tools.}
    \begin{tabular}{lccccc}
        \toprule
        Model & \multicolumn{3}{c}{Planning} & \multicolumn{2}{c}{Recognition} \\ 
        \cmidrule(lr){2-4} \cmidrule(lr){5-6}
        &VR & P & R & Exact & GPTScore \\
        \midrule
        Ours w/ specialized tools &0.9960& \topone{0.9105}& \topone{0.9728}& \topone{0.7563}& \topone{0.7848}  \\
        Ours w/o specialized tools &\topone{0.9990}& 0.8556& 0.9448& 0.5332& 0.6092  \\
        \bottomrule
    \end{tabular}
    \label{tab:abl_tool}
\end{table*}

\subsection{Quantitative Analysis}

We evaluate the proposed method on the \datasetname{} dataset. As summarized in Table.~\ref{tab:task_model_metrics}, we compare the capability of our agent-based method with existing models across three critical aspects: planning, perception, and recognition. Notably, traditional object detection and semantic segmentation methods only produce perception results while the VQA methods focus solely on producing recognition results. In contrast, our method addresses all three dimensions, demonstrating its comprehensive capabilities.

\subsubsection{Planning performance}

Planning is the most important stage in \taskname{} as the proper usage of tools is crucial to the interpretation results. Among all related tasks, our framework is the only one that produces explicit planning outputs. The quality of the planning stage is evaluated with valid rate, precision, and recall. A high valid rate indicates that the proposed agent produces a reliable planning format instead of outputting corrupted text. 

With GPT-4o-mini as the LLM backend, our agent-based approach consistently produces valid plans in over 99\% of cases. The planning achieves a precision of 90.51\% and a recall of 97.24\%, showing the robustness and effectiveness of the agent-based framework.

\subsubsection{Perception performance}

As our approach utilizes state-of-the-art perception models as a part of its tool library, the accuracy of the final answer is linked to the performance of these models. Therefore, we conduct experiments with different detection and segmentation models on \datasetname{} dataset and select PSPNet~\cite{pspnet} and GroundingDINO~\cite{groundingdino} as the most accurate and reliable tools.

To evaluate the perception performance, we compute metrics directly on the raw segmentation and detection labels for more comprehensive results. In our proposed framework, the perception models are scheduled by the planning module, and the perception outputs are directly fed to the recognition model. Therefore, our proposed method inherits the accuracy of the segmentation model and the detection model used to build the tool library, achieving a 68.28\% mIoU for segmentation and an 80.10\% average precision for object detection.

\subsubsection{Recognition performance}

The recognition metrics demonstrate the advantage of our agent-based method over end-to-end LLM-based VQA methods. With the utilization of explicit perception models as tools, the agent produces more accurate answers, offering intermediate outputs such as image segmentation and object detection to enhance decision-making processes. We compare the agent-based method with two baseline methods, VisualGLM~\cite{ding2021cogview_visualglm} and GeoChat~\cite{kuckreja2023geochatgroundedlargevisionlanguage}. Since VisualGLM is primarily developed for natural scenes instead of remote sensing scenarios, we conduct extra tuning with LoRA~\cite{hu2022lora} to align it with the recognition requirements of the \datasetname{} dataset. Our approach achieves 73.74\% exact-match accuracy and 80.90\% GPTScore, which is superior compared to both tuned VisualGLM and GeoChat. 

To better understand the advantage of the agent-based framework, we inspect the planning and recognition accuracy for each individual request type. As shown in Table. \ref{tab:per_type}, end-to-end VQA models tend to make mistakes on numerical requests such as counting and area calculations as they do not explicitly produce perception results of the image. Equipped with dedicated perception modules, our agent-based approach exhibits strong accuracy for numerical answers. Our approach automatically determines the perception models to use and then selects specialized tools to perform counting and area calculation tasks accurately. For rescue path-finding requests, the advantage is more obvious as the task is easy for specialized tools but difficult to model by end-to-end neural networks.

\subsection{Ablation Studies}

\textbf{LLM Backend:} The proposed agent-based framework is designed to be compatible with different LLM backends as long as the LLM is tuned to follow input instructions. To better understand the effect of different LLM backends, we experiment with several state-of-the-art instruction-following LLMs: GPT 4o mini, Deepseek 7B, Qwen1.5 7B, Qwen2.5 7B, and Llama 3.1 8B. The LLMs are inserted into the framework and act as the text generator of both the planning agent and the summarizing agent. Results in Table. \ref{tab:abl_llm} shows that the capability of the LLM backends has a strong correlation to the planning quality of the agent and, therefore impacts the subsequent perception and recognition. The closed-source LLM, GPT 4o mini, exhibits the best capability while the best open-source LLM for the task is Qwen2.5 7B. The planning precision of Qwen2.5 7B is slightly higher than that of GPT 4o mini, but the recall is significantly lower, resulting in reduced overall accuracy. Therefore, we conclude that a strong LLM Backend is crucial in the proposed agent-based framework.

\textbf{Specialized Tools:} In our proposed method, we employ specialized tools in the recognition module to help analyze the perception result. These tools cover counting, area calculation, and pathfinding. To investigate the contribution of these tools, we perform experiments without specialized tools as shown in Table.~\ref{tab:abl_tool}. As specialized tools are removed, we observe a significant drop in recognition accuracy. Both exact match score and GPTScore drop by more than 15\%, indicating that specialized tools play an important role in \taskname{}.

\textbf{Performance Boundaries:} Powered by pretrained LLMs, the proposed agent-based method is able to generalize toward more complex requests. As shown in Fig.~\ref{fig_ood} (a), the agent can handle compound questions that requires the perception of multiple categories. The request asks to count the number of buildings in different damage levels and the planning agent successfully gives a reasonable plan that uses the counting tool to process each damage level. Fig.~\ref{fig_ood} (b) shows a request that requires advanced reasoning. The agent performs both segmentation and detection to obtain the total area and the number of buildings that are destroyed, and utilizes these numbers to calculate the correct answer.

\section{Conclusion} \label{s_conclusion} 

In this paper, we propose a novel adaptive disaster interpretation task to address the problem of extra human intervention and lack of accuracy in current disaster interpretation pipelines. On top of that, we produce the \datasetname{} dataset, the first dataset for \taskname{}, that integrates planning, perception, and recognition of disaster scenarios, serving as a robust benchmark. We propose the first autonomous agent-based method with specialized tools to solve the challenging disaster interpretation task. Experiment results prove the effectiveness of the agent-based framework on \taskname{} task, showing an accuracy improvement of more than 9\% compared to existing VQA methods. However, experiments have also indicated that the capability of the LLM backend can be crucial to the overall results. Our future work will be focused on improving the robustness of the whole framework and further expanding the variety of the dataset.


\bibliographystyle{IEEEtran}
\bibliography{refs}

\vfill

\end{document}